# Development of a Robotic System for Automatic Wheel Removal and Fitting

Gideon Gbenga Oladipupo
*Engineering and Physical Sciences School*
*Heriot Watt University*
Edinburgh, United Kingdom
ggo1@hw.ac.uk;
hgrv86@gmail.com

*Abstract* —**This paper discusses the image processing and computer vision algorithms for real time detection and tracking of a sample wheel of a vehicle. During the manual tyre changing process, spinal and other muscular injuries are common and even more serious injuries have been recorded when occasionally, tyres fail (burst) during this process. It, therefore, follows that the introduction of a robotic system to take over this process would be a welcome development. This work discusses various useful applicable algorithms, Circular Hough Transform (CHT) as well as Continuously adaptive mean shift (Camshift) and provides some of the software solutions which can be deployed with a robotic mechanical arm to make the task of tyre changing faster, safer and more efficient. Image acquisition and software to accurately detect and classify specific objects of interest were implemented successfully, outcomes were discussed and areas for further studies suggested.**

*Keywords—Camshift, Automatic, Wheel, Robotic, CHT.*

## I. INTRODUCTION

Global demand for the automation of tasks especially those with inherent risks and hazard has been on the rise in the last few decades. Fundamentally, automation exists to substitute tasks undertaken by human labour with machines, with the aim of increasing quality and quantity of output at a minimal cost [1]. The advent of automation has numerous applications in various fields of life ranging from industries, military, medicine and space exploration amongst others. Particularly, automobile industries have witnessed a high statistics of automation applications and employment [1]. Fitting the wheel of a typical large mining vehicle requires a minimum of two workers using heavy machines to carry out the operation which could be hazardous. The employment of robotic platforms for this operation offers a better option in terms of safety, efficiency and speed [2]. Robotic systems have a substantial level of autonomy enabled by the computer vision system to carry out wheel removing and fitting operations [3]. Techniques for image processing and object detection by machine vision would be evaluated with a view to developing a more robust approach for autonomous operations in an industrial setting. The rest of the paper is organized into the review of previous works and theoretical, proposed methodology as well as the results and discussion.

## II. PREVIOUS WORKS AND THEORETICAL REVIEW

### A. Previous Tracking Methods

Numerous approaches were previously researched [4-7] on conveyor tracking of object or line tracking. Conveyor tracking according to [5] essentially revolves around a pre-programmed robot performing desired operations on a workpiece as it moves on conveyor belt. Some of the previous tracking approaches include use of optical flow, where fibre-optics connected to a remote camera was used to track the moving objects, and adaptive approach for controlling the manipulator with the assumption that the object velocity is constant over a vision sampling period [6,7]

### B. Fundamentals of Image Processing

Image processing according to [8] reinforced by [9] is the process of performing computational operations on images with the intent of improving the perception of image content for human viewers and to also extract its interesting features for numerical measurement. This is significant for robots as it would be investigated in this research. Image processing essentially has to do with digital image mostly arranged in an array consisting of sampled image values. Each of the indexed elements in the array are otherwise known as picture elements (pixel) and they are pointers for objection detection and recognition. The key stages of image processing techniques as shown in Figure 1 are image acquisition, image enhancement, image restoration and morphological processing. Others are segmentation, object recognition and representation as well as description [9].

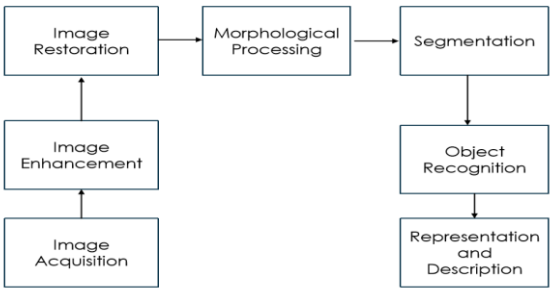

*Figure 1: Image Processing stages [8,9].*

### C. Circular Hough Transform

Fundamentally, Hough Transform (HT) implements a voting process that maps image edge points into manifolds in an appropriately defined parameter space[10-14]. The three-stage process for circle detection and the voting pattern are shown in Figure 3 and Figure respectively. Peaks in the space correspond to the parameters of detected curves. Circular Hough Transform (CHT), is designed to find a circle characterized by a center point ($x_0$, $y_0$) and a radius r [13]. Mathematically, CHT is represented as

$$(x - x_{centre})^2 + (y - y_{centre})^2 = r^2 \ldots\ldots\ldots\ldots\ldots\ldots\ldots 1$$

where ( $x_{centre}$, $y_{centre}$ ) is the centre of the circle, and *r* is the radius of the circle. It is obvious that the equation has 3 parameters, hence 3D accumulator is needed for the hough transform. This is however ineffective due to the computational cost. To determine a circle for instance, it is necessary to accumulate votes in the three-dimensional parameter space ($x_0$, $y_0$, r). The challenge of computational inefficiency is however addressed in Open Computer Vision (OpenCV) using Hough gradient.

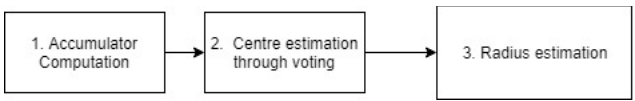

*Figure 2: Three steps implementation of CHT Algorithm.*

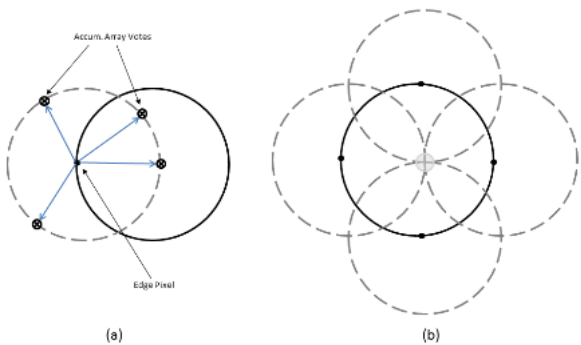

*Figure 3: Typical CHT Voting Pattern [10].*

### D. Circular Hough Transform in OpenCV

The implementation of CHT in OpenCV helps to mitigate the computational cost associated with CHT in terms of memory requirements and speed of the algorithm[15,16]. This is achieved through a method called Hough gradient method.

In this method, the image is passed through an edge detection phase using canny edge detector - cvCanny(), in this case, the local gradient of every nonzero point in the edge image is considered [15] . The gradient is computed by first computing the first-order Sobel *x*- and *y*-derivatives via cvSobel(). With this gradient, every point along the line indicated by the slope – from a specified minimum to a specified maximum distance – is incremented in the accumulator. Similarly, the location of every one of these nonzero pixels in the edge is noted [15]. The candidate centres are then selected from points in the 2D accumulator that are both above some given threshold and larger than all their immediate neighbours. These candidate centres are sorted in a descending order of their accumulator value, such that the centres with the most supporting pixels appear first [15,16]. For each centre, all the nonzero pixels are sorted according to their distance from the centre. Adopting the smallest distances to the maximum radius, a single radius is selected that is best supported by the nonzero pixels. A centre is therefore kept if it has sufficient support from the nonzero pixels in the edge image and if it is a sufficient distance from any previously selected centre [15, 17].

The OpenCV function that implements Hough circle detection is called HoughCirlces(). The following are the input parameters:

- Grayscale image, on which Canny edge detection is applied.
- Inverse ratio of the accumulator resolution to image resolution, for circle centres detection. For instance, if it is set to 2, the accumulator is half the size of the image.
- Minimum distance between the centres of detected circles. If this parameter is too small, multiple neighbour circles may be falsely detected in addition to the true one. Similarly, if it is too large, some circles may be missed. Hence, this parameter has to be carefully adjusted to detect true circle while the false circles are rejected.
- Higher threshold of the Canny edge detector for pre-processing the input image, the lower threshold is set to half of the higher threshold.
- The accumulator threshold for the circle centres at the detection stage, the smaller it is, the more false circles may be detected.

Minimum and maximum circle radius, can also be used to filter out noisy small and false large circles [15, 18].

### E. Review of Object Tracking Algorithms

The information gathered from detection algorithm could be used as a basis for the tracking of object in an image. Several tracking algorithms have previously been implemented. The choice of algorithms however depends on the object features such as colour, motion as well as edges to be tracked and the environmental conditions [19,20]. Similarly, the performance of tracking algorithms usually depends on the measure characterizing the similarity or dissimilarity between the two subsequent images/video frames [20]. Some of the tracking algorithms are outlined subsequently:

- *Background Subtraction*

Essentially, background subtraction identifies moving objects from the portion of a video frame that differs significantly from a background model [21]. This algorithm is based on the assumption that the colours of a moving object is different those in the background [24]. Additionally, background subtraction algorithm assumes that the camera is fixed firmly in location with a static noise-free background [22,23]. However, this would not the case in real life scenarios as changes in illumination and other environmental factors affects the background and camera position [22]. The four major steps in a background subtraction algorithm as shown in Figure 4 are pre-processing, background modelling, foreground detection, and data validation [21].

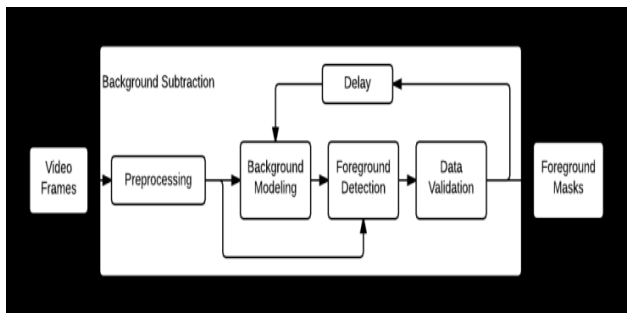

*Figure 4: Flow diagram of a generic background subtraction algorithm[21]*

- *Mean Shift Tracking Algorithms*

The mean-shift algorithm is a robust method of finding local extrema in the density distribution of a data set [25]. It has been proposed as method for cluster analysis [26-29]. [26] gave a mathematical representation where data represents a finite set $S$ embedded in the n-dimensional Euclidean space, $X$ while $K$ is a flat kernel that is the characteristic function of the λ-ball in $X$,

$$K(x) = \begin{cases} 1 = if\ ||x|| \leq \lambda \\ 0 = if\ ||x|| > \lambda \end{cases} \ldots\ldots\ldots\ldots\ldots 1$$

The sample mean at $x \in X$ is

$$m(x) = \frac{\sum_{s \in S} K(s-x)s}{\sum_{s \in S} K(s-x)} \ldots\ldots\ldots\ldots\ldots 2$$

The difference $m(x) - x$ is called mean shift in [27] while the repeated movement of data points to the sample means is called the mean-shift algorithm [27, 28]. In each iteration of the algorithm, $s \leftarrow m(s)$ is performed for all $s \in S$ simultaneously. The implementation of mean-shift algorithm is as follows:

- Choose a search window:
  - Its initial action.
  - Its type (uniform, polynomial, exponential, or Gaussian).
  - Its shape (symmetric or skewed, possibly rotated, rounded or rectangular).
  - Its size (extent at which it rolls off or is cut off).
- Compute the window's (possibly weighted) centre of mass.
- Centre the window at the centre of mass.
- Return to second step until the windows stops moving.

III. PROPOSED METHODOLOGY

*Camshift Tracking Algorithms*

In this paper, Camshift algorithm as implemented in OpenCV was used to track the hub of the wheel sample in different lighting conditions and at different distances from the camera. Camshift algorithm closely related to mean-shift algorithm [30,31]. It differs from the mean-shift in that the algorithm adjusts the window size and pixel distribution pattern during tracking. While mean-shift uses static probability distributions which are not updated during tracking, Camshift improved on this, by using continuously adaptive probability distributions. This could ensure that the target's probability distribution is recomputed in every frame[32-34].

Essentially, Camshift climbs the gradient of a back-projected probability distribution which is computed from a re-scaled colour histogram in finding the nearest peak within a search window [32, 35]. Hence, the mean location of a target is found by computing zeroth, first and second order image moments:

$$M_{00} = \sum_x \sum_y P(x,y) \ldots\ldots\ldots\ldots\ldots\ldots\ldots 3$$

$$M_{10} = \sum_x \sum_y xP(x,y);\ M_{01} = \sum_x \sum_y yP(x,y)\ldots 4$$

$$M_{20} = \sum_x \sum_y x^2 P(x,y);\ M_{02} = \sum_x \sum_y y^2 P(x,y) \ldots 5$$

Where $P(x,y) = h(I(x,y))$ is back projected probability distribution at position $x, y$ within the search window $I(x,y)$ that is computed from the histogram $h$ of $I$ [32]. The target object's mean position can then be computed with

$$x_c = \frac{M_{10}}{M_{00}};\ y_c = \frac{M_{01}}{M_{00}} \ldots\ldots\ldots\ldots 6$$

while its aspect ratio

$$ratio = \frac{M_{20}/x_c^2}{M_{02}/y_c^2};\ \ldots\ldots\ldots\ldots 7$$

is used for updating the search window with

$$width = 2M_{00}\,.\,ratio;\ height = 2M_{00}/ratio \ldots 8$$

The implementation of Camshift algorithm as stated in [36] are highlighted in the following steps:
- Set the region of interest (ROI) of the probability distribution image to the entire image.
- Select an initial location of the Mean Shift search window. The selected location is the target distribution to be tracked.

- o Calculate a colour probability distribution of the region centred at the Mean Shift search window.
- o Iterate Mean Shift algorithm to find the centroid of the probability image. Store the zeroth moment (distribution area) and centroid location.
- o For the following frame, centre the search window at the mean location found in previous step and set the window size to a function of the zeroth moment. Go to Step 3.

The workflow for the implementation of Camshift tracking algorithm is as shown in Figure 5.

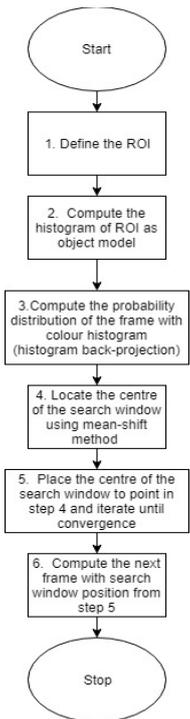

*Figure 5: Camshift Algorithm workflow*

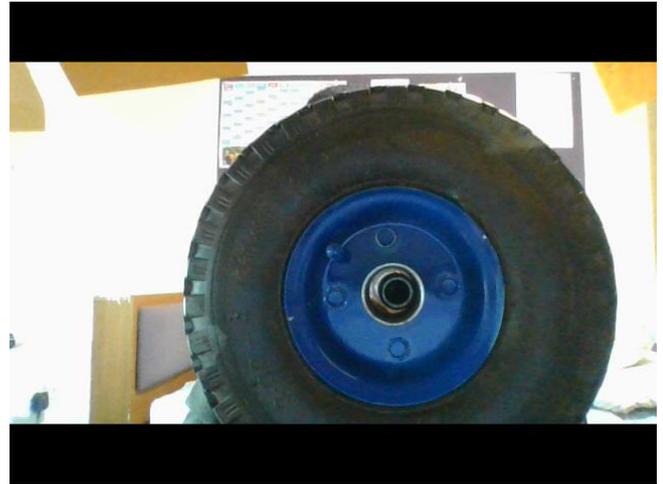

*Figure 6: Original wheel tyre captured using webcam with OpenCV on python.*

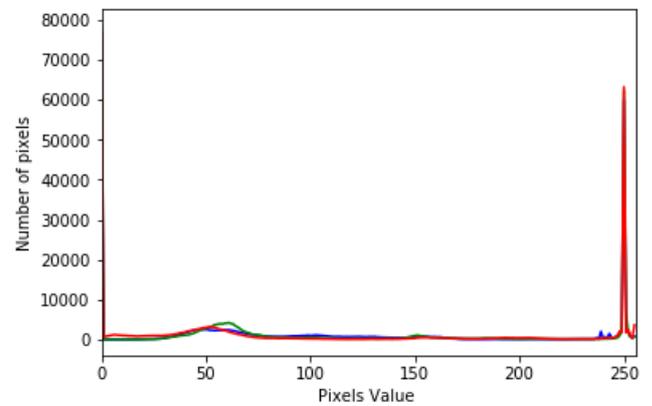

*Figure 7: BGR Colour distribution of sample tyre wheel.*

IV. RESULTS AND DISCUSSION

A. *Image Acquisition*

Image of the sample tyre wheel was acquired in OpenCV on python programming language using the webcam from the laptop. The original sample wheel as captured using the laptop webcam is shown in Figure 6 while the histogram distribution of the BGR colours is shown in Figure 7. On applying the Hough circle algorithm on the tyre wheel, the circular wheel hub as well as the main tyre wheel were captured as shown in Figure 8 and Figure 9 respectively. This was attained by adjusting the parameters to obtain the desired circular object size.

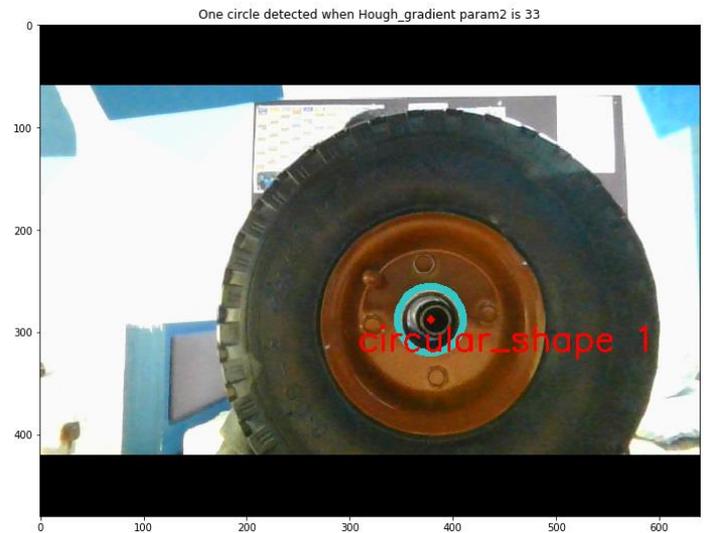

*Figure 8: Detected wheel hub using the Hough Circle Algorithm on OpenCV.*

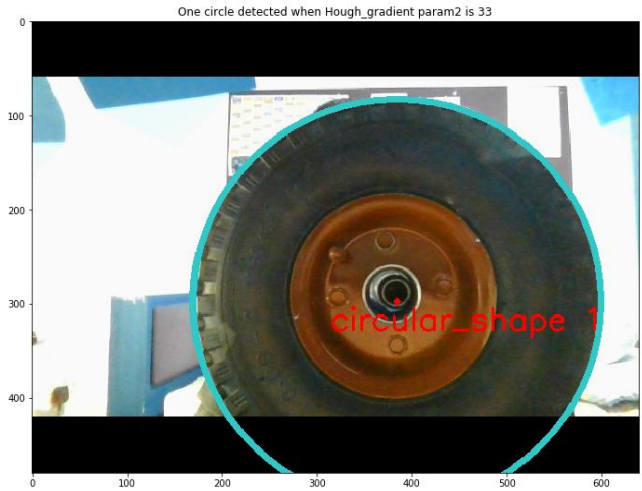

*Figure 9: Detected Tyre Wheel using the Hough Circle Algorithm on OpenCV.*

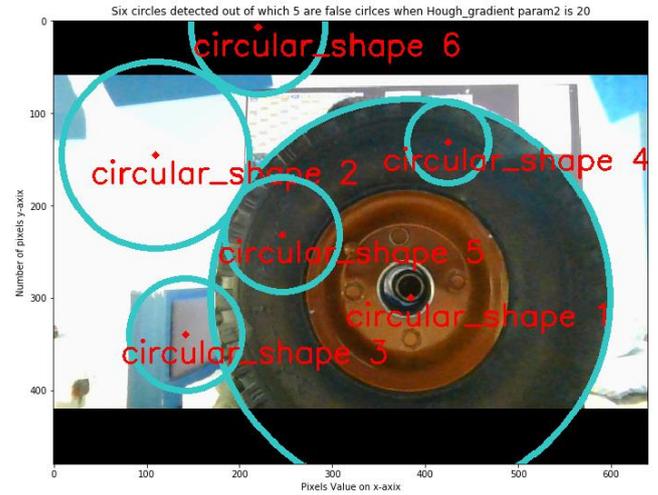

*Figure 10: Detection of 5 false circles in addition to the main tyre wheel.*

### B. Performance Analysis of CHT Algorithm

The performance of CHT algorithm is significantly affected by the parameters tuning of the algorithm. The parameters as previously highlighted include the thresholds for the canny edge detector and the accumulator space for the pixels. Others are minimum and maximum radius of the expected circle as well as the inverse ratio of the accumulator to image resolution and the minimum distance between centres of the detected circles. Each of the parameters shown in Table 1 has a default values which could be adjusted to get the desired result. Using the parameter values in the third column (column c) of Table 1 and applying them in the CHT algorithm on the wheel sample image returned 5 false circles in addition to the main tyre which was correctly detected. This agreed with assertion of [15, 18], which states that if the minimum distance between the centres of detected circles is too small, multiple neighbour circles may be falsely detected in addition to the true one. Similarly, if it is too large, some circles may be missed. As seen in Figure 10, the wheel hub was not detected using this setting. This explains the importance of correct setting of the parameter values to get the desired result.

*Table 1: Parameters setting leading to false circle detection*

| Parameters | Default Values | Values used |
|---|---|---|
| (a) | (b) | (c) |
| Inverse ratio of accumulator to image resolution (Dp) | 0.1-1.0 | 0.8 |
| Minimum distance between centres of detected circles | 1-10000 | 150 |
| Canny edge threshold (Param_1) | 200 | 50 |
| Accumulator threshold (Param_2) | 100 | 20 |
| Minimum radius of the expected circle | 0 | 0 |
| Maximum radius of the expected circle | 0 | 0 |

*Table 2: Parameter setting for the correct detection of wheel hub*

| Parameters | Default Values | Values used |
|---|---|---|
| Inverse ratio of accumulator to image resolution (Dp) | 0.1-1.0 | 0.1 |
| Minimum distance between centres of detected circles | 1-10000 | 18 |
| Canny edge threshold (Param_1) | 200 | 50 |
| Accumulator threshold (Param_2) | 100 | 33 |
| Minimum radius of the expected circle | 0 | 0 |
| Maximum radius of the expected circle | 0 | 50 |

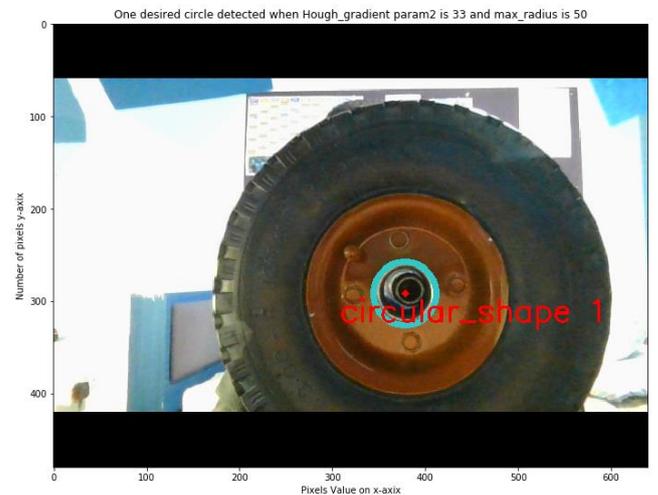

*Figure 11: Detected wheel hub with radius of 34 pixels and centre coordinate of [378 292] pixels.*

Similarly, on careful tweaking of the parameters as shown in

Table 2 , the wheel hub was detected. In this case, the minimum distance between centres of detected was reduced to 18 pixels as against 150 pixels as reflected in Table 1 while the accumulator threshold was increase to 33 pixels as against 20 pixels as indicated in the previous table. Also, the maximum radius was selected to be 50 pixels, thereby preventing the likelihood of false circles detection. The detected wheel hub is shown in Figure 11. This outcome correlates with the stated principles in [15] on the correct adjustment of the parameters for circle detection.

Furthermore, the maximum radius was tweaked to its default values of 0 while other parameters values used for the detection of the wheel hub were left unchanged. The used parameters are shown in Table 3 while the detected tyre wheel is shown in Figure 12. The centre coordinates for the detected tyre is [384 300] pixels while the radius is 215 pixels.

*Table 3: Parameter setting for the detection of tyre*

| Parameters | Default Values | Values used |
|---|---|---|
| Inverse ratio of accumulator to image resolution (Dp) | 0.1-1.0 | 0.1 |
| Minimum distance between centres of detected circles | 1-10000 | 18 |
| Canny edge threshold (Param_1) | 200 | 50 |
| Accumulator threshold (Param_2) | 100 | 33 |
| Minimum radius of the expected circle | 0 | 0 |
| Maximum radius of the expected circle | 0 | 0 |

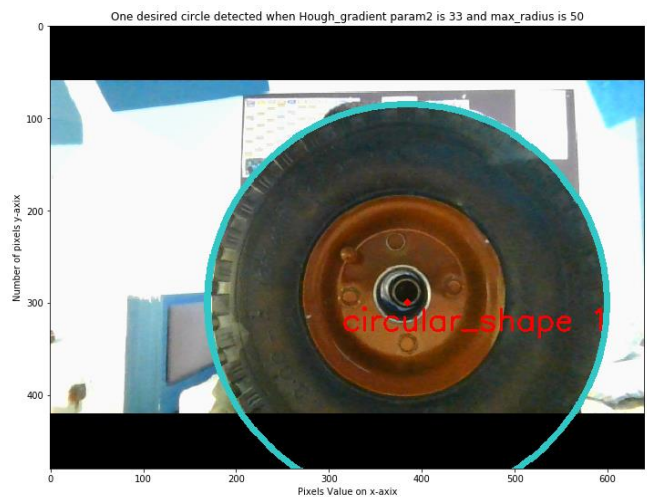

*Figure 12: Detected tyre with radius of 215 pixels and centre coordinate of [384 300] pixel.*

### C. Performance Analysis of Camshift Algorithm

The Camshift tracking algorithm was employed for the tracking of the wheel hub after its detection from CHT algorithm. The hub centre coordinates as well as the radius was used to defined the needed region of interest, which, essentially was a tool for target localisation [38]. In this case, the captured region of interest was analysed and its 3D colour channel histogram is shown in Figure 13 while its 1D colour channel is shown in *Figure 14*. The conversion to 1D colour channel was to facilitate easy recognition and tracking as the machine is seeing one colour at this instance, thereby, reducing the vision complexity associated with multiple colours channel.

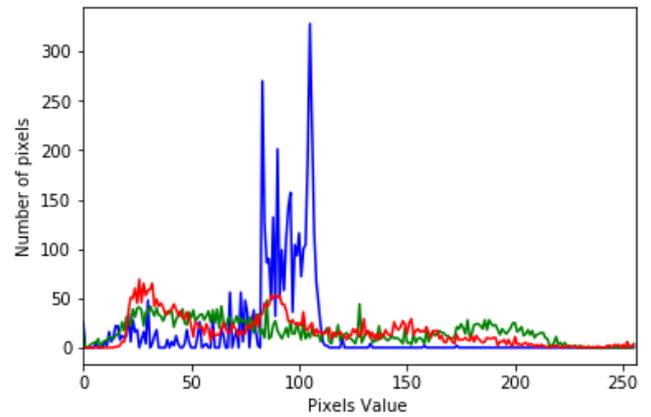

*Figure 13: The two peaks represent wheel centre coordinates at (292,378) in 3D colour channel*

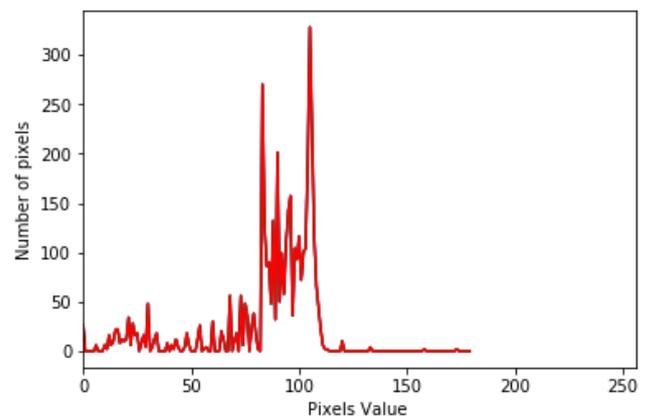

*Figure 14: The two peaks represent wheel centre coordinates at (292,378) in 1D colour channel.*

The tracking effectiveness and correctness of the Camshift algorithm was tested under different room lighting conditions ranging from lightly dim room to moderately lit room and well-lit room. Additionally, the algorithm was tested by varying the distance from about 1m to 2m and then to 3m.

Figure 15 depicts the performance of the algorithm in 3 different lighting conditions at about 1m away for the laptop webcam. The red box represents the defined tracking window which is derived from the ROI of the wheel image. In order to achieve real time detection, the ROI was defined to track only the wheel hub as against the whole tyre image. This is consistent with the idea of [39] where the HSV colour conversion was performed only on the defined ROI, thereby reducing the processing time. This in turn facilitates real time computation and tracking of the desired object region. In this case, the algorithm could effectively track the wheel hub under the different lighting conditions investigated at about 1m away from the camera.

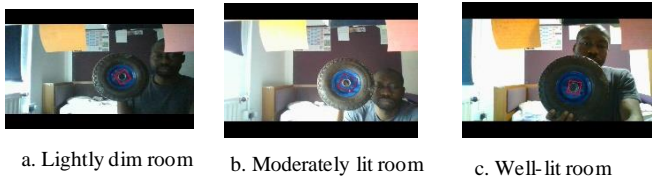

*Figure 15: Performance of Camshift Algorithm at about 1m from the Laptop webcam under different lighting condition*

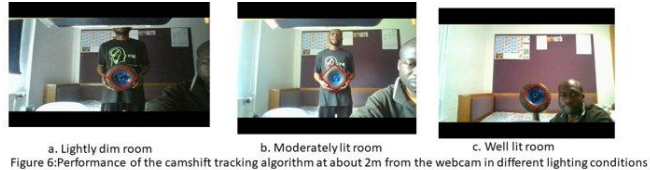

*Figure 16: Performance of Camshift Algorithm at about 2m from the Laptop webcam under different lighting condition.*

Furthermore, the tyre wheel was moved farther away to about 2m to observe its performance as shown in Figure 16. In this case, the tracking window resizes and readjusts itself to track the entire wheel. This outcome agrees with the investigation earlier carried out in [40] where the tracking window was seen to readjust itself as the object to be tracked moves away from the camera. Essentially, this ability of the Camshift algorithm to adjust itself is in line with its underlying principle to adapt itself continuously with the change in its environment. The target object to be tracked must however remain within the field of view of the camera for this to prove true.

The final test was also carried out by moving the tyre wheel to 3m away from the webcam under the different room lighting conditions as shown in Figure 17. The algorithm performed similarly to when the tyre wheel was at 2m away from the webcam, but this time around, the tracking window has readjusted itself to accommodate the change in the distance. This affirms the robustness of the Camshift algorithm in the face of changing environment condition as stated in [37] where the algorithm used with Kalman filter. Kalman filter was used to predict the velocity and position of the moving target. Kalman filter was not used in this research but other important aspect of Camshift algorithm was seen to agree with the [37].

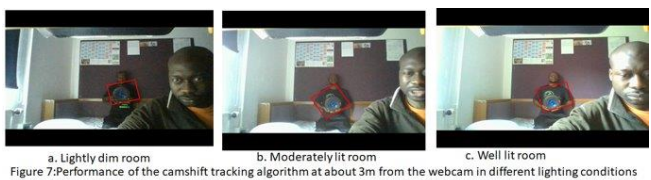

*Figure 17: Performance of Camshift Algorithm at about 3m from the Laptop webcam under different lighting condition.*

CONCLUSION

This paper has sought to contribute to the furtherance of deploying fully automated robotic systems for automated wheel changing and has proffered solutions to some challenges currently being faced by existing systems. It's hoped that some of the discoveries and suggestions made here can be used to improve existing robotic wheel changing systems. Some of the suggested areas where further studies could be carried out include observing the performance of Camshift algorithm under different light intensities as well as the employment of Kinect sensor version 2 in OpenCV.